\documentclass{article}

\usepackage[preprint]{neurips_2021}

\usepackage[utf8]{inputenc} 
\usepackage[T1]{fontenc}    
\usepackage{hyperref}       
\usepackage{url}            
\usepackage{booktabs}       
\usepackage{amsfonts}       
\usepackage{nicefrac}       
\usepackage{microtype}      
\usepackage{xcolor}         
\usepackage{latexsym}
\usepackage{graphics}
\usepackage{epsfig,amssymb,amsmath,graphicx,amsfonts,bm,amsthm}

\newcommand{\beqan}{\begin{eqnarray*}}
\newcommand{\eeqan}{\end{eqnarray*}}
\newcommand{\beqa}{\begin{eqnarray}}
\newcommand{\eeqa}{\end{eqnarray}}
\newcommand{\beq}{\begin{equation}}
\newcommand{\eeq}{\end{equation}}

\newtheorem{theorem}{Theorem}
\newtheorem{lemma}{Lemma}

\newtheorem{prop}{\textbf{Proposition}}

\newtheorem{definition}{\textbf{Definition}}

\title{On Linear Separability under Linear Compression with Applications to Hard Support Vector Machine}

\author{%
  Paul McVay \\
  Department of Electrical Engineering\\
  Texas A\&M University\\
  College Station, TX 77843 \\
  \texttt{pmcvay@tamu.edu} \\
  \And
  Tie Liu \\
  Department of Electrical Engineering\\
  Texas A\&M University\\
  College Station, TX 77843 \\
  \texttt{tieliu@tamu.edu} \\
  \And
  Krishna Narayanan \\
  Department of Electrical Engineering\\
  Texas A\&M University\\
  College Station, TX 77843 \\
  \texttt{krn@tamu.edu} \\
}

\begin{document}

\maketitle

\begin{abstract}
This paper investigates the theoretical problem of maintaining linear separability of the data-generating distribution under linear compression. While it has been long known that linear separability may be maintained by linear transformations that approximately preserve the inner products between the domain points, the limit to which the inner products are preserved in order to maintain linear separability was unknown. In this paper, we show that linear separability is maintained as long as the distortion of the inner products is smaller than the squared margin of the original data-generating distribution. The proof is mainly based on the geometry of hard support vector machines (SVM) extended from the finite set of training examples to the (possibly) infinite domain of the data-generating distribution. As applications, we derive bounds on the (i) compression length of random sub-Gaussian matrices; and (ii) generalization error for compressive learning with hard-SVM.

\end{abstract}

\section{Introduction}\label{sec:intro}
Compressed learning is the process of running machine learning algorithms on data that have been compressed. Compressed learning was originally studied under the context of kernel support vector machines (SVM). It was first observed that the sample complexity of SVM after transforming the data into a high-dimensional kernel space does not grow proportionally with the dimension of the kernel space. One explanation introduced by \citet{balcan2006kernels} used the idea of compressed learning. That is, the benefit of kernel SVM can be thought of as a {\em two-step} feature mapping transformation. First, the original data are mapped to a high-dimensional space where the features are linearly separable with a high margin. Next, the high-dimensional data are compressed into a low-dimensional space where the margin is preserved. The second mapping step, compressed learning, was also studied in \cite{blum2005random,calderbank2009compressed,shi2012margin}.

The more contemporary motivation for studying compressed learning is based on the fact that the size of datasets are growing very rapidly today. For example, the ODP dataset with 422,712 features for each data point in each of the 105,033 classes would require more than 160 GB to store a standard logistic regression model \cite{daume2017logarithmic}. Reducing the number of features through compression has the potential of substantially reducing the model size. Note that while a hashing scheme such as \citet{weinberger2009feature} can mitigate the run-time dependency on the feature size, most standard machine learning algorithms do not have this capability and would run significantly faster with a reduced feature size.

The need for reducing the data transmission cost in server/node configurations can be found in many applications such as traffic analysis based on packet timing and sizes. With encryption becoming a common practice, content-based traffic analysis becomes nearly impossible. Researchers have been exploring traffic analysis based on packet timing and sizes to detect cyber attacks. In the basic setup for this line of research, many network relay nodes send packet statistics to a central server. The central server uses machine learning techniques to identify the anonymized users and attacks. Ideally, these are identified in real-time. The data that the nodes send to the central server must be compressed based on the channel capacity. Although some compression techniques allow for the recovery of the original signal, the real-time constraint of the analysis makes it impractical for the server to reconstruct each signal, thus forcing the server to run the machine learning algorithm directly on the compressed data \cite{nasr2017compressive}.

Although there are many signal compression techniques, we will restrict our attention in this paper to the techniques of using a matrix to linearly project the data onto a lower-dimensional space and study the performance loss incurred by running binary classification algorithms directly on the compressed data. Our focus is the simplest setting where the original data is in a high-dimensional space but is known {\em a priori} to be linearly separable with high margin. Projecting data onto lower dimensions will introduce distortions to the {\em geometry} of the data, which will in turn lead to reductions of the margin and eventually loss of the linear separability. Our goal is to understand to what extent the data can be compressed while still maintaining the linear separability, and our main result is a fairly intuitively satisfying relationship between the preservation of the margin and the preservation of the geometry: Linear separability can be maintained if the distortion of the {\em inner products} is smaller than the squared margin of the original data. 

Note that previous works \cite{balcan2006kernels,blum2005random,shi2012margin}
 have also
studied margin preservation under linear projection. However, there
are two key differences between the previous works and the current
work. First, while the previous works focused on maintaining linear
separability for a {\em finite} set of data examples, the focus of
this paper is on maintaining linear separability at the {\em
  distribution} level, i.e., the data-generating distribution is
assumed to be linearly separable (with certain margin) before
compression and must remain linearly separable after compression. The
fact that the domain of the data-generating distribution may be {\em
  infinite} requires significant extra care during the technical
development. Second, while the previous works focused on the
performance of {\em random} projection matrices, this paper focuses on
understanding the relationship between preservation of margin and
preservation of geometry. Such a relationship is more fundamental in
that it can be used to evaluate the performance of any projection
matrix including random projection matrices.

The rest of the paper is organized as follows. Next in Section~\ref{sec:main} we provide a formal statement of the problem and present our main results on the relationship between preservation of margin and preservation of geometry under linear compression of data. In section~\ref{sec:rand} we derive bounds on the compression length of random sub-Gaussian matrices by leveraging the main results of section~\ref{sec:main}. As another application, in Section~\ref{sec:gen} we derive a generalization bound for compressed learning with hard-SVM, which illustrates how the bound on the generalization error of hard-SVM changes after compression.

\section{Problem setup and main results}\label{sec:main}
Consider a binary classification problem for which the data-generating distribution is given by $\mu=\mu_\mathsf{x}\mu_{\mathsf{y}|\mathsf{x}}$. Let $\mathcal{X}^+$ be the {\em support} of $\mu_{\mathsf{x}|\mathsf{y}=+1}$, i.e., 
\begin{align}
\mathcal{X}^+:=\left\{x\in\mathbb{R}^n: \mbox{$\mu(\mathcal{A}\times\{+1\})>0$ for any $\mathcal{A}\subseteq \mathbb{R}^n$ such that $\mathcal{A}$ is open and $x\in \mathcal{A}$}\right\}.
\end{align}
Similarly, let $\mathcal{X}^-$ be the support of $\mu_{\mathsf{x}|\mathsf{y}=-1}$. Further let 
\begin{align}
\mathcal{X}: =\mathcal{X}^+\cup\mathcal{X}^{-} \quad \mbox{and} \quad \mathcal{Z}: =\left(\mathcal{X}^+\times\{+1\}\right)\cup\left(\mathcal{X}^-\times\{-1\}\right).
\end{align} 
Note that $\mathcal{Z}$ is measurable because both $\mathcal{X}^+$ and $\mathcal{X}^-$ are closed and $\mu(\mathcal{Z})=1$. 

We say that the data-generating distribution $\mu$ is {\em linearly separable} if there exists a hyperplane in $\mathbb{R}^n$ indexed by $(w_0,b_0)$ such that 
\begin{align}
\mathbb{P}_{(\mathsf{x},\mathsf{y})\sim \mu}\left[\mathsf{y}\left(\langle w_0,\mathsf{x}\rangle+b_0\right)\geq 1\right]=1.
\end{align}
Following \cite{shalev2014understanding}, the {\em margin} of the hyperplane is given by $1/\|w_0\|$.

Let $Q \in \mathbb{R}^{m\times n}$ be an $m$-by-$n$ matrix of real entries. Assuming that $\mu$ is linearly separable, we say that $\mu$ remains {\em linearly separable under the compression} of $Q$ if there exists a hyperplane in $\mathbb{R}^m$ indexed by $(w',b')$ such that 
\begin{align}
\mathbb{P}_{(\mathsf{x},\mathsf{y})\sim \mu}\left[\mathsf{y}\left(\langle w',Q\mathsf{x}\rangle+b'\right)\geq 1\right]=1.
\end{align}
Intuitively, $\mu$ remains linearly separable under the compression of $Q$ if $Q$ approximately preservers the {\em geometry} of $\mathcal{X}$, the support of $\mu_\mathsf{x}$. We use the following definition to describe the preservation of geometry over a given set.

\begin{definition}[Inner product preserving linear transformation]
Let $\mathcal{A}$ be a subset of $\mathbb{R}^n$ and let $Q \in \mathbb{R}^{m\times n}$ be an $m$-by-$n$ matrix of real entries. We say that the matrix $Q$ is {\em $\eta$-inner-product-preserving} over $\mathcal{A}$ if
\begin{align}
\sup_{x,x'\in\mathcal{A}}\left|\langle Qx,Qx'\rangle-\langle x,x'\rangle\right| \leq \eta,\label{eq:ip}
\end{align}
where $\eta$ is the {\em distortion} of the inner products over $\mathcal{A}$ introduced by $Q$.
\end{definition}

Assume that $\mu$ is linearly separable by a hyperplane in $\mathbb{R}^n$ indexed by $(w_0,b_0)$. We emphasize here that $(w_0,b_0)$ is prior knowledge, which in general cannot be estimated from the data samples generated from $\mu$. Let $Q \in \mathbb{R}^{m\times n}$ be $\eta$-inner-product-preserving over $\mathcal{X}$, the support of $\mu$. Our goal is to understand the limit of $\eta$ for which $\mu$ would remain linearly separable under the compression of $Q$. Our main result is summarized in the following theorem.

\begin{theorem}\label{thm:main}
Assume that the data-generating distribution $\mu$ is linearly separable by the hyperplane in $\mathbb{R}^n$ indexed by $(w_0,b_0)$. Let $Q\in\mathbb{R}^{m\times n}$ be $\eta$-inner-product-preserving over $\mathcal{X}$, where $\mathcal{X}$ is the support of $\mu_{\mathsf{x}}$. Then, $\mu$ remains linearly separable under the compression of $Q$ if $\eta<\frac{1}{\|w_0\|^2}$. 
\end{theorem}

In words, the data-generating distribution $\mu$ remains linearly separable under the compression of $Q$ if the distortion of the inner products over the support of $\mu$ introduced by $Q$ is smaller than the squared margin of $\mu$ based on the prior knowledge. While the above result seems to be rather natural and intuitive, our proof is quite nontrivial. More specifically, in order to show that  $\mu$ remains linearly separable under the compression of $Q$, we need to construct a hyperplane in $\mathbb{R}^m$ that can separate the induced distribution $Q\mu$. One may naturally consider constructions based on $w_0$. However, $w_0$ is prior knowledge and in practice can be quite arbitrary. In particular, $w_0$ can behave very differently than the vectors from $\mathcal{X}$ under the compression of $Q$, which is very problematic for our proof. To overcome this issue, our proof relies on the following notion of compatibility and the existence of a compatible solution. 

\begin{definition}[Compatible vectors] 
Let $\mathcal{A}$ be a subset of $\mathbb{R}^n$ and let $Q \in \mathbb{R}^{m\times n}$ be $\eta$-inner-product-preserving over $\mathcal{A}$. We say that a vector $w\in\mathbb{R}^m$ is {\em $c$-compatible} with $\mathcal{A}$ under $Q$ if
\begin{align}
\sup_{x\in \mathcal{A}}\left|\langle Qw,Qx\rangle-\langle w,x\rangle\right| \leq c\eta,
\end{align}
where $c$ measures the extent to which $w$ is compatible with the vectors from $\mathcal{A}$ under the compression of $Q$. Note that we can set $c=1$ for any $w\in\mathcal{A}$.
\end{definition}

The following proposition is a key result to the proof of Theorem~\ref{thm:main}.

\begin{prop}[Existence of a compatible solution]\label{prop:comp}
Assume that the data-generating distribution $\mu$ is linearly separable by the hyperplane in $\mathbb{R}^n$ indexed by $(w_0,b_0)$. Let $Q \in \mathbb{R}^{m\times n}$ be $\eta$-inner-product-preserving over $\mathcal{X}$, where $\mathcal{X}$ is the support of $\mu_{\mathsf{x}}$. Then there exists $(w^*,b^*)$ such that: (i) $\mu$ is linearly separable by the hyperplane indexed by $(w^*,b^*)$; (ii) $\|w^*\| \le \|w_0\|$; and (iii) $w^*$ is $\|w^*\|^2$-compatible with $\mathcal{X}$ under $Q$.
\end{prop}

The construction of the compatible solution $(w^*,b^*)$ is mainly based on the geometry of hard support vector machines (SVM) \cite{bennett2000duality}, extended from the finite set of training examples to the (possibly) infinite support of the data-generating distribution $\mu$. The proof of Proposition~\ref{prop:comp} and Theorem~\ref{thm:main} can be found in Sections~\ref{pf:prop-comp} and \ref{pf:thm-main}, respectively.

\section{Bounds on the compression length of random sub-Gaussian matrices}\label{sec:rand}
In this section, we derive bounds on the compression length of random sub-Gaussian matrices by leveraging the result of Theorem~\ref{thm:main}. Our strategy is to first relate the notion of preservation of inner products with the more traditional notion of preservation of squared distances and then leverage the concentration results on preservation of squared distances.

\begin{definition}[squared distance preserving linear transform]
Let $\mathcal{A}$ be a subset of $\mathbb{R}^n$ and let $Q \in \mathbb{R}^{m\times n}$ be an $m$-by-$n$ matrix of real entries. We say that the matrix $Q$ is {\em $\eta$-squared-distance-preserving} over $\mathcal{A}$ if
\begin{align}
\sup_{x,x'\in\mathcal{A}}\left|\|Qx-Qx'\|^2-\|x-x'\|^2\right| \leq \eta,\label{eq:sd}
\end{align}
where $\eta$ is the {\em distortion} of the squared distances over $\mathcal{A}$ introduced by $Q$.
\end{definition}

The notions of inner product squared distance preservations are closely related to each other. In particular, note that 
\begin{align*}
\|Qx-Qx'\|^2=\langle Qx,Qx\rangle+\langle Qx',Qx'\rangle-2\langle Qx,Qx'\rangle.
\end{align*}
Therefore, if for some $x,x'\in\mathcal{A}$ we have
\begin{align*}
\left|\langle Qx,Qx'\rangle-\langle x,x'\rangle\right| \leq \eta,
\end{align*}
we also have
\begin{align*}
\left|\|Qx-Qx'\|^2-\|x,x'\|^2\right| \leq 4\eta.
\end{align*}
That is, if $Q$ is $\eta$-inner-product-preserving over $\mathcal{A}$, it is also $4\eta$-squared-distance-preserving over $\mathcal{A}$.

The converse results require additional assumptions on the set $\mathcal{A}$. Let us first consider the case where the vectors from $\mathcal{A}$ are {\em sparse}. More specifically, let $\mathcal{A}$ be a subset of $\mathbb{R}^n$ such that $\|x\|\leq R$ and $\|x\|_0 \leq s$ for all $x\in \mathcal{A}$. By triangle inequality, we have $\|x-x'\|\leq 2R$ and $\|x-x'\|_0 \leq 2s$ for all $x,x'\in \mathcal{A}$. Then, for any let $Q \in \mathbb{R}^{m\times n}$ we have
\begin{align*}
\sup_{x,x'\in\mathcal{A}}\left|\|Qx-Qx'\|^2-\|x-x'\|^2\right| \leq \sup_{x,x'\in\mathcal{A}}\delta_{2s}\|x-x'\|^2\leq 4\delta_{2s}R^2,
\end{align*}
where $\delta_{2s}$ is the {\em $2s$-restricted isometric constant} of
$Q$ \cite{foucart2013mathematical}, i.e., $Q$ is
$4\delta_{2s}R^2$-squared-distance-preserving over $\mathcal{A}$. It
can be shown that $Q$ is also inner-product-preserving over
$\mathcal{A}$, as stated in the following proposition. The proof of
the proposition can be found in the apendix in Section~\ref{pf:prop-sparse}.

\begin{prop}\label{prop:sparse}
Let $\mathcal{A}$ be a subset of $\mathbb{R}^n$ such that $\|x\|\leq R$ and $\|x\|_0 \leq s$ for all $x\in \mathcal{A}$, and let $Q \in \mathbb{R}^{m\times n}$ be an $m$-by-$n$ matrix of real entries. Then $Q$ is $\delta_{2s}R^2$-inner-product-preserving over $\mathcal{A}$, where $\delta_{2s}$ is the $2s$-restricted isometric constant of $Q$.
\end{prop}

For a general set of vectors $\mathcal{A}$, note that
\begin{align*}
\langle x,x'\rangle =\frac{\|x-0\|^2+\|x'-0\|^2-\|x-x'\|^2}{2}.
\end{align*}
Assuming that $0 \in \mathcal{A}$, we have the following simple converse result.

\begin{prop}\label{prop:gen}
Let $\mathcal{A}$ be a subset of $\mathbb{R}^n$ such that $0 \in \mathcal{A}$, and let $Q \in \mathbb{R}^{m\times n}$ be an $m$-by-$n$ matrix of real entries. If $Q$ is $\eta$-squared-distance-preserving over $\mathcal{A}$, it is also $\frac{3}{2}\eta$-inner-product-preserving over $\mathcal{A}$. 
\end{prop}

The converse results discussed above allow us to establish bounds on the compression length from the concentration properties of random sub-Gaussian matrices in in terms of squared distance preservations. For sparse vectors, we quote the following result \cite[Theorem~9.6]{foucart2013mathematical} on the restricted isometric constant of a scaled sub-Gaussian random matrix.

\begin{theorem}[\cite{foucart2013mathematical}]
\label{thm:sparse}
Let $Q\in\mathbb{R}^{m\times n}$ be an $m\times n$ random matrix with independent, isotropic, sub-Gaussian rows with the same sub-Gaussian norm. If
\begin{align*}
m \geq C\delta^{-2}\left(s\ln(en/s)+\ln(2\varepsilon^{-1})\right),
\end{align*}
the restricted isometric constant $\delta_s$ of $\frac{1}{\sqrt{m}}Q$ satisfies $\delta_s \leq \delta$ with probability at least $1-\varepsilon$, where $C$ only depends on the sub-Gaussian norm of the rows of $Q$.
\end{theorem}

Combining Proposition~\ref{prop:sparse} and Theorem~\ref{thm:sparse} immediately leads to the following result.

\begin{theorem}
Let $\mu$ be a data-generating distribution such that: (i) $\mu$ is linearly separable by the hyperplane in $\mathbb{R}^n$ indexed by $(w_0,b_0)$, and (ii) $\|x\|\leq R$ and $\|x\|_0 \leq s$ for all $x\in \mathcal{X}$, where $\mathcal{X}$ is the support of $\mu_{\mathsf{x}}$. Let $Q \in \mathbb{R}^{m\times n}$ be an $m\times n$ random matrix with independent, isotropic, sub-Gaussian rows with the same sub-Gaussian norm. If 
\begin{align}
m > CR^4\|w_0\|^4\left(2s\ln(en/(2s))+\ln(2\varepsilon^{-1})\right),
\end{align}
$\mu$ remains linearly separable under the compression of $\frac{1}{\sqrt{m}}Q$ with probability at least $1-\varepsilon$, where $C$ only depends on the sub-Gaussian norm of the rows of $Q$.
\end{theorem}

For general vectors, we quote the following Johnson-Lindenstrauss lemma for infinite sets \cite[Proposition~9.3.2]{vershynin2018high}.

\begin{theorem}[\cite{vershynin2018high}]
\label{thm:JL}
Let $Q\in\mathbb{R}^{m\times n}$ be an $m\times n$ random matrix with independent, isotropic, sub-Gaussian rows with the same sub-Gaussian norm, and let $\mathcal{A}$ be a subset of $\mathbb{R}^n$. With probability at least $1-\varepsilon$, $\frac{1}{\sqrt{m}}Q$ is 
\begin{align*}
\frac{K^2\left[w(\mathcal{A})+\ln(2\varepsilon^{-1})r(\mathcal{A})\right]^2+2\sqrt{m}K\left[w(\mathcal{A})+\ln(2\varepsilon^{-1})r(\mathcal{A})\right]r(\mathcal{A})}{m}
\end{align*}
squared-distance-preserving over $\mathcal{A}$, where $r(\mathcal{A}):=\sup_{x\in\mathcal{A}}\|x\|$ is the {\em radius} of $\mathcal{A}$, $w(\mathcal{A})$ is the {\em Gaussian width} of $\mathcal{A}$, and $K$ is a constant that only depends on the sub-Gaussian norm of the rows of $Q$.
\end{theorem}

Combining Proposition~\ref{prop:gen} and Theorem~\ref{thm:JL} immediately leads to the following result.

\begin{theorem}
Let $\mu$ be a data-generating distribution such that: (i) $\mu$ is linearly separable by the hyperplane in $\mathbb{R}^n$ indexed by $(w_0,b_0)$, and (ii) $0\in\mathcal{X}$ and $\|x\|\leq R$ for all $x\in \mathcal{X}$, where $\mathcal{X}$ is the support of $\mu_{\mathsf{x}}$. Let $Q \in \mathbb{R}^{m\times n}$ be an $m\times n$ random matrix with independent, isotropic, sub-Gaussian rows with the same sub-Gaussian norm. If the compression length $m$ satisfies
\begin{align}
\frac{3}{2}\cdot\frac{K^2\left[w(\mathcal{A})+\ln(2\varepsilon^{-1})R\right]^2+2\sqrt{m}K\left[w(\mathcal{A})+\ln(2\varepsilon^{-1})R\right]R}{m}<\frac{1}{\|w_0\|^2},
\end{align}
$\mu$ remains linearly separable under the compression of $\frac{1}{\sqrt{m}}Q$ with probability at least $1-\varepsilon$, where $w(\mathcal{A})$ is the Gaussian width of $\mathcal{A}$ and $K$ is a constant that only depends on the sub-Gaussian norm of the rows of $Q$.
\end{theorem}

\section{Generalization bound for compressed learning with hard-SVM}\label{sec:gen}
We now apply our previous results to derive a generalization bound for
compressed learning with hard-SVM. We note that this application
requires the knowledge of separability that is preserved at the
{\em distribution} level. Recall that if the data-generating distribution $\mu$ is linearly separable and has a bounded domain, i.e., $\mathbb{P}_{(\mathsf{x},\mathsf{y})\sim \mu}\left[\|\mathsf{x}\|\leq R\right]=1$, let $(w_\mathcal{S}, b_\mathcal{S})$ be the output of hard-SVM over a training dataset $\mathcal{S}$. Then, with probability at least $1-\delta$, the true error of $(w_\mathcal{S}, b_\mathcal{S})$ can be bounded from above as 
\begin{align}
\mathbb{P}_{(\mathsf{x}, \mathsf{y}) \sim \mu}\left[\mathsf{y}\left(\langle w_\mathcal{S},\mathsf{x}\rangle+b_\mathcal{S}\right) \leq 0\right] \le L(\|w_\mathcal{S}\|),
\end{align}
where
\begin{align}
L(z) := \frac{8Rz+2+\sqrt{\ln(4\delta^{-1}\log_2z)}}{\sqrt{|\mathcal{S}|}}.
\end{align}

The following result shows how the bound on the generalization error of hard-SVM changes after compression.

\begin{theorem} \label{thm:gen}
Assume that the data-generating distribution $\mu$ is linearly separable by the hyperplane in $\mathbb{R}^n$ indexed by $(w_0,b_0)$ and has a bounded domain, i.e., $\mathbb{P}_{(\mathsf{x},\mathsf{y})\sim \mu}\left[\|\mathsf{x}\|\leq R\right]=1$. Let $Q\in\mathbb{R}^{m\times n}$ be $\eta$-inner-product-preserving over $\mathcal{X}$, where $\mathcal{X}$ is the support of $\mu_{\mathsf{x}}$ and $\eta<\frac{1}{\|w_0\|^2}$. Then, $\mu$ remains linearly separable under the compression of $Q$. Furthermore, let $(w_\mathcal{S},b_\mathcal{S})$ and $(\bar{w}_\mathcal{S}, \bar{b}_\mathcal{S})$ be the output of hard-SVM over a train dataset $\mathcal{S}$ before and after compression, respectively. Then with probability at least $1-\delta$, the true error of $(\bar{w}_\mathcal{S}, \bar{b}_\mathcal{S})$ can be bounded from above as:
\begin{align}
\mathbb{P}_{(\mathsf{x}, \mathsf{y}) \sim \mu}\left[\mathsf{y}\left(\langle \bar{w}_\mathcal{S},Q\mathsf{x}\rangle+\bar{b}_\mathcal{S}\right) \leq 0\right] \le \frac{L(\|w_\mathcal{S}\|)}{\sqrt{1-\eta\|w_\mathcal{S}\|^2}},
\end{align} 
where $L(\|w_\mathcal{S}\|)$ the generalization bound of $(w_\mathcal{S},b_\mathcal{S})$ before compression.
\end{theorem}

This results shows that we can bound the generalization error of the compressed hard-SVM solution without actually compressing and computing the hard-SVM solution. Thus, we can compute the ideal compression length for a given accuracy requirement without iteratively recomputing the compressed dataset and solution. The proof of the theorem is omitted from the paper due to the space limitation.

\section{Proofs}
\subsection{Proof of Proposition~\ref{prop:comp}}\label{pf:prop-comp}
We shall prove the results of the proposition through the following
sequence of lemmas. The first three lemmas are simply stated here while
the proofs are in the appendix.  

\begin{lemma}[Support]
Let $\mu$ be the data-generating distribution.
\begin{itemize}
\item[i)] If $\mathbb{P}_{(\mathsf{x},\mathsf{y})\sim \mu}\left[\|\mathsf{x}\|\leq R\right]=1$, then $\|x\|\leq R$ for all $x\in\mathcal{X}$.
\item[ii)] If $\mathbb{P}_{(\mathsf{x},\mathsf{y})\sim \mu}\left[\|\mathsf{x}\|_0\leq s\right]=1$, then $\|x\|_0\leq s$ for all $x\in\mathcal{X}$.
\item[iii)] If $\mathbb{P}_{(\mathsf{x},\mathsf{y})\sim \mu}\left[\mathsf{y}\left(\langle w_0,\mathsf{x}\rangle+b_0\right)\geq 1\right]=1$, then $y\left(\langle w_0,x\rangle+b_0\right)\geq 1$ for all $(x,y)\in\mathcal{Z}$.
\end{itemize}
\label{lm:support}
\end{lemma}

{\em Proof:} Appendix

Let 
\begin{align}
\delta:=\inf\left\{\|x'-x''\|: x'\in co\left(\mathcal{X}^+\right),\;x''\in co\left(\mathcal{X}^-\right)\right\}.
\end{align}

\begin{lemma}[Strict separation between $co\left(\mathcal{X}^+\right)$ and $co\left(\mathcal{X}^-\right)$]
  $\delta\geq\frac{2}{\|w_0\|}>0$.
  \label{lm:strict-separation}
\end{lemma}

{\em Proof:} Appendix

\begin{lemma}[Achievability of the minimum separation]
There exists $\bar{x}'\in cl\left(co\left(\mathcal{X}^+\right)\right)$
and $\bar{x}''\in cl\left(co\left(\mathcal{X}^+\right)\right)$ such
that $\|\bar{x}'-\bar{x}''\|=\delta$.
\label{lm:achieve-min-sep}
\end{lemma}

{\em Proof:} Appendix

Let 
\begin{align}
w^* &:= \frac{2(\bar{x}'-\bar{x}'')}{\|\bar{x}'-\bar{x}''\|^2}\\
\mbox{and} \quad b^* & := 1-\frac{2\langle \bar{x}'-\bar{x}'',
  \bar{x}'\rangle}{\|\bar{x}'-\bar{x}''\|^2}.
\label{eqn:geometric-svm}
\end{align}

\begin{lemma}
$\|w^*\|\leq \|w_0\|$.
\end{lemma}

{\em Proof:} By the definition of $w^*$ and the facts that $\|\bar{x}'-\bar{x}''\|=\delta$ and $\delta \geq \frac{2}{\|w_0\|}$, we have
$$\|w^*\|=\frac{2}{\|\bar{x}'-\bar{x}''\|}=\frac{2}{\delta}\leq\frac{2}{2/\|w_0\|}=\|w_0\|.$$

\begin{lemma}
$\mathbb{P}_{(\mathsf{x},\mathsf{y})\sim \mu}\left[\mathsf{y}\left(\langle w^*,\mathsf{x}\rangle+b^*\right)\geq 1\right]=1$.
\end{lemma}

{\em Proof:} Note that by construction $\mu(\mathcal{Z})=1$, so it suffices to show that $y\left(\langle w^*,x\rangle+b^*\right)\geq 1$ for all $(x,y)\in\mathcal{Z}$. To show that $y\left(\langle w^*,x\rangle+b^*\right)\geq 1$ for all $(x,y)\in\mathcal{Z}$, we need to show that $\langle w^*,x\rangle+b^*\geq 1$ for any $x\in\mathcal{X}^+$ and $\langle w^*,x\rangle+b^*\leq -1$ for any $x\in\mathcal{X}^-$. 

Let us first consider the case where $x\in\mathcal{X}^+$. Note that for any $x\in\mathbb{R}^n$, we have
\begin{align*}
\langle w^*,x\rangle+b^* &=\left\langle \frac{2(\bar{x}'-\bar{x}'')}{\|\bar{x}'-\bar{x}''\|^2},x\right\rangle+\left(1-\frac{2\langle \bar{x}'-\bar{x}'', \bar{x}'\rangle}{\|\bar{x}'-\bar{x}''\|^2}\right)\\
&=1+\frac{2\langle \bar{x}'-\bar{x}'',x-\bar{x}'\rangle}{\|\bar{x}'-\bar{x}''\|^2}.
\end{align*}
Therefore, to show that $\langle w^*,x\rangle+b^*\geq 1$ for any $x\in\mathcal{X}^+$, it suffices to show that $\langle \bar{x}'-\bar{x}'',x-\bar{x}'\rangle\geq0$ for any $x\in\mathcal{X}^+$.

To show that $\langle \bar{x}'-\bar{x}'',x-\bar{x}'\rangle\geq0$ for any $x\in\mathcal{X}^+$, let $(x'_n:n\in\mathbb{N})$ be a sequence in $co\left(\mathcal{X}^+\right)$ that converges to $\bar{x}'$ and let $(x''_n:n\in\mathbb{N})$ be a sequence in $co\left(\mathcal{X}^-\right)$ that converges to $\bar{x}''$. By the convexity of $co\left(\mathcal{X}^+\right)$, we have $\lambda x+(1-\lambda)x'_n \in co\left(\mathcal{X}^+\right)$ for any $n\in\mathbb{N}$ and any $\lambda\in(0,1)$. It follows that
$$\|\left(\lambda x+(1-\lambda)x'_n\right)-x''_n\| \geq \delta= \|\bar{x}'-\bar{x}''\|$$
for any $n\in\mathbb{N}$. Let $n \rightarrow \infty$ and we have by the continuity of the norm function $\|\cdot\|$ that
$$\|\left(\lambda x+(1-\lambda)\bar{x}'\right)-\bar{x}''\|=\|\lambda(x-\bar{x}')+(\bar{x}'-\bar{x}'')\| \geq \|\bar{x}'-\bar{x}''\|,$$
which can be equivalently written as
$$\lambda^2\|x-\bar{x}'\|^2+2\lambda\langle x-\bar{x}',\bar{x}'-\bar{x}''\rangle \geq 0$$
or
$$\langle x-\bar{x}',\bar{x}'-\bar{x}''\rangle \geq -\frac{\lambda\|x-\bar{x}'\|^2}{2}$$
for any $\lambda\in(0,1)$. Letting $\lambda\downarrow 0$ completes the proof that $\langle \bar{x}'-\bar{x}'',x-\bar{x}'\rangle\geq0$ for any $x\in\mathcal{X}^+$.

Next, let us consider the case where $x\in\mathcal{X}^-$. Note that for any $x\in\mathbb{R}^n$, we have
\begin{align*}
\langle w^*,x\rangle+b^* &=1+\frac{2\langle \bar{x}'-\bar{x}'',x-\bar{x}'\rangle}{\|\bar{x}'-\bar{x}''\|^2}\\
&=-1+\frac{2\langle \bar{x}'-\bar{x}'',\bar{x}'-\bar{x}''\rangle}{\|\bar{x}'-\bar{x}''\|^2}+\frac{2\langle \bar{x}'-\bar{x}'',x-\bar{x}'\rangle}{\|\bar{x}'-\bar{x}''\|^2}\\
&=-1+\frac{2\langle \bar{x}'-\bar{x}'',x-\bar{x}''\rangle}{\|x-\bar{x}''\|^2}.
\end{align*}
Therefore, to show that $\langle w^*,x\rangle+b^*\leq 1$ for any $x\in\mathcal{X}^-$, it suffices to show that $\langle \bar{x}'-\bar{x}'',x-\bar{x}''\rangle\leq0$ for any $x\in\mathcal{X}^-$.

To show that $\langle \bar{x}'-\bar{x}'',x-\bar{x}''\rangle\geq0$ for any $x\in\mathcal{X}^-$, again let $(x'_n:n\in\mathbb{N})$ be a sequence in $co\left(\mathcal{X}^+\right)$ that converges to $\bar{x}'$ and let $(x''_n:n\in\mathbb{N})$ be a sequence in $co\left(\mathcal{X}^-\right)$ that converges to $\bar{x}''$. By the convexity of $co\left(\mathcal{X}^-\right)$, we have $\lambda x+(1-\lambda)x''_n \in co\left(\mathcal{X}^+\right)$ for any $n\in\mathbb{N}$ and any $\lambda\in(0,1)$. It follows that
$$\|x'_n-\left(\lambda x+(1-\lambda)x''_n\right)\| \geq \delta= \|\bar{x}'-\bar{x}''\|$$
for any $n\in\mathbb{N}$. Let $n \rightarrow \infty$ and we have by the continuity of the norm function $\|\cdot\|$ that
$$\|\bar{x}'-\left(\lambda x+(1-\lambda)\bar{x}''\right)\|=\|(\bar{x}'-\bar{x}'')-\lambda(x-\bar{x}'')\| \geq \|\bar{x}'-\bar{x}''\|,$$
which can be equivalently written as
$$\lambda^2\|x-\bar{x}''\|^2-2\lambda\langle \bar{x}'-\bar{x}'',x-\bar{x}''\rangle \geq 0$$
or
$$\langle \bar{x}'-\bar{x}'',x-\bar{x}'\rangle \leq \frac{\lambda\|x-\bar{x}''\|^2}{2}$$
for any $\lambda\in(0,1)$. Letting $\lambda\downarrow 0$ completes the proof that $\langle \bar{x}'-\bar{x}'',\bar{x}'-\bar{x}''\rangle\leq0$ for any $x\in\mathcal{X}^-$. 

Combining both cases from above completes the proof of the lemma. \hfill $\blacksquare$

\begin{lemma}
If $Q \in \mathbb{R}^{m\times n}$ is $\eta$-inner-product-preserving over $\mathcal{X}=\mathcal{X}^+\cup\mathcal{X}^-$, then $w^*$ is $\|w^*\|^2$-compatible with $\mathcal{X}$ under $Q$.
\end{lemma}

{\em Proof:} To show that $w^*$ is $\|w^*\|^2$-compatible with $\mathcal{X}$ under $Q$, let $(x'_n:n\in\mathbb{N})$ be a sequence in $co\left(\mathcal{X}^+\right)$ that converges to $\bar{x}'$ and let $(x''_n:n\in\mathbb{N})$ be a sequence in $co\left(\mathcal{X}^-\right)$ that converges to $\bar{x}''$. For each $n\in\mathbb{N}$, by Carath\'{e}odory's theorem we can write $x'_n=\sum_{i=1}^{n+1}\alpha_{n,i}x'_{n,i}$ for some $x'_{n,i} \in \mathcal{X}^+$, $i\in[n+1]$, and $\alpha_{n,i} \geq 0$, $i\in[n+1]$, such that $\sum_{i=1}^{n+1}\alpha_{n,i}=1$. Similarly, for each $n\in\mathbb{N}$ we can write $x''_n=\sum_{i=1}^{n+1}\beta_{n,i}x''_{n,i}$ for some $x''_{n,i} \in \mathcal{X}^-$, $i\in[n+1]$, and $\beta_{n,i} \geq 0$, $i\in[n+1]$, such that $\sum_{i=1}^{n+1}\beta_{n,i}=1$.

Fix $x\in\mathcal{X}$. Note that for any $n\in\mathbb{N}$, we have
\begin{align*}
\langle Qx'_n,Qx\rangle &= \left\langle Q\left(\sum_{i=1}^{n+1}\alpha_{n,i}x'_{n,i}\right),Qx\right\rangle= \sum_{i=1}^{n+1}\alpha_{n,i}\langle Qx'_{n,i},Qx\rangle.
\end{align*}
By assumption, $Q$ is $\eta$-inner-product-preserving over $\mathcal{X}$, so we have
$$\langle x'_{n,i},x\rangle-\eta \leq \langle Qx'_{n,i},Qx\rangle \leq \langle x'_{n,i},x\rangle+\eta.$$
It follows that
$$\sum_{i=1}^{n+1}\alpha_{n,i}\left(\langle x'_{n,i},x\rangle-\eta\right) \leq \langle Qx'_n,Qx\rangle \leq \sum_{i=1}^{n+1}\alpha_{n,i}\left(\langle x'_{n,i},x\rangle+\eta\right).$$
Further note that
\begin{align*}
\sum_{i=1}^{n+1}\alpha_{n,i}\langle x'_{n,i},x\rangle=\left\langle \sum_{i=1}^{n+1}\alpha_{n,i}x'_{n,i},x\right\rangle=\langle x'_n,x\rangle.
\end{align*}
We thus have
$$\langle x'_n,x\rangle-\eta \leq \langle Qx'_n,Qx\rangle \leq \langle x'_n,x\rangle+\eta.$$
Let $n\rightarrow \infty$. By the continuity of the linear transform $Q\cdot$ and the inner product $\langle\cdot,x\rangle$, we have
$$\langle \bar{x}',x\rangle-\eta \leq \langle Q\bar{x}',Qx\rangle \leq \langle \bar{x}',x\rangle+\eta.$$

Similarly, it can be shown that 
$$\langle \bar{x}'',x\rangle-\eta \leq \langle Q\bar{x}'',Qx\rangle \leq \langle \bar{x}'',x\rangle+\eta.$$
Combining the above two results gives
$$\langle \bar{x}'-\bar{x}'',x\rangle-2\eta \leq \langle Q(\bar{x}'-\bar{x}''),Qx\rangle \leq \langle \bar{x}'-\bar{x}'',x\rangle+2\eta.$$
Multiplying each side of the inequalities by $\frac{2}{\|\bar{x}'-\bar{x}''\|^2}$, we have
$$\left\langle\frac{2(\bar{x}'-\bar{x}'')}{\|\bar{x}'-\bar{x}''\|^2},x\right\rangle-\frac{4\eta}{\|\bar{x}'-\bar{x}''\|^2} \leq \left\langle Q\frac{2(\bar{x}'-\bar{x}'')}{\|\bar{x}'-\bar{x}''\|^2},Qx\right\rangle \leq \left\langle\frac{2(\bar{x}'-\bar{x}'')}{\|\bar{x}'-\bar{x}''\|^2},x\right\rangle+\frac{4\eta}{\|\bar{x}'-\bar{x}''\|^2}.$$
Using the facts that $w^*=\frac{2(\bar{x}'-\bar{x}'')}{\|\bar{x}'-\bar{x}''\|^2}$ and $\|w^*\|=\frac{2}{\|\bar{x}'-\bar{x}''\|}$, we 
rewrite the above inequality as
$$\langle w^*,x\rangle-\eta\|w^*\|^2 \leq \langle Qw^*,Qx\rangle \leq \langle w^*,x\rangle+\eta\|w^*\|^2$$
or equivalently
$$\left|\langle Qw^*,Qx\rangle-\langle w^*,x\rangle\right| \leq \eta\|w^*\|^2.$$
Since the above inequality holds for all $x\in\mathcal{X}$, we conclude that $w^*$ is $\|w^*\|^2$-compatible with $\mathcal{X}$ under $Q$. \hfill $\blacksquare$

We have thus completed the proof of Proposition~\ref{prop:comp}.

\subsection{Proof of Theorem~\ref{thm:main}}\label{pf:thm-main}
By assumption, $\mu$ is linearly separable by the hyperplane in $\mathbb{R}^n$ indexed by $(w_0,b_0)$ and $Q$ is $\eta$-inner-product-preserving over $\mathcal{X}$. By Proposition~1, there exists another hyperplane in $\mathbb{R}^d$ indexed by $(w^*,b^*)$ such that 
\begin{align}
\mathbb{P}_{(\mathsf{x},\mathsf{y})\sim \mu}\left[\mathsf{y}\left(\langle w^*,\mathsf{x}\rangle+b^*\right)\geq 1\right]=1,\label{eq:T0}
\end{align}
$\|w^*\| \le \|w_0\|$, and $w^*$ is $\|w^*\|^2$-compatible with $\mathcal{X}$ under $Q$, i.e.,
\begin{align}
\left|\langle Qw^*,Qx\rangle-\langle w^*,x\rangle\right| \leq \eta\|w^*\|^2, \quad \forall \; x\in \mathcal{X}.
\end{align}
Note that by Lemma~3, \eqref{eq:T0} implies that $y\left(\langle w^*,x\rangle+b^*\right)\geq 1$ for all $(x,y)\in\mathcal{Z}$.
 
Next, we show that for any $(x,y)\in\mathcal{Z}$, we have
\begin{align}
y\left(\langle Qw^*,Qx\rangle+b^*\right) \geq 1-\eta\|w^*\|^2.\label{eq:T1}
\end{align}
Let us consider the cases where $y=+1$ and $y=-1$, separately. If $y=+1$, we have $\langle w^*,x\rangle+b^*\geq 1$. It follows that
\begin{align}
y\left(\langle Qw^*,Qx\rangle+b^*\right) &= \langle Qw^*,Qx\rangle+b^*\nonumber\\
& \geq \langle w^*,x\rangle-\eta\|w^*\|^2+b^*\nonumber\\
& \geq (1-b^*)-\eta\|w^*\|^2+b^*\nonumber\\
& = 1-\eta\|w^*\|^2.\label{eq:T2}
\end{align}
Similarly, if $y=-1$, we have $\langle w^*,x\rangle+b^*\leq -1$. It follows that
\begin{align}
y\left(\langle Qw^*,Qx\rangle+b^*\right) &= -\langle Qw^*,Qx\rangle-b^*\nonumber\\
& \geq -\langle w^*,x\rangle-\eta\|w^*\|^2-b^*\nonumber\\
& \geq (1+b^*)-\eta\|w^*\|^2-b^*\nonumber\\
& = 1-\eta\|w^*\|^2.\label{eq:T3}
\end{align}
Combining \eqref{eq:T2} and \eqref{eq:T3} completes the proof of \eqref{eq:T1}.

By assumption, we have $\eta<\frac{1}{\|w_0\|^2}$ and hence 
\begin{align*}
1-\eta\|w^*\|^2>1-\frac{\|w^*\|^2}{\|w_0\|^2} \geq 0
\end{align*}
where the last inequality follows from the fact that $\|w^*\| \le \|w_0\|$. We can thus rewrite \eqref{eq:T0} as
\begin{align}
y\left(\left\langle \frac{Qw^*}{1-\eta\|w^*\|^2},Qx\right\rangle+\frac{b^*}{1-\delta\|w^*\|^2}\right) \geq 1
\end{align}
which immediately implies that the data-generating distribution $\mu$, when compressed by $Q$, can be linearly separated by the hyperplane in $\mathbb{R}^m$ indexed by 
$$\left(\frac{Qw^*}{1-\delta\|w^*\|^2},\frac{b^*}{1-\delta\|w^*\|^2}\right).$$
This completes the proof of Theorem~\ref{thm:main}.

{
\small
\bibliography{citations}
}

\newpage
\appendix

\section{Appendix}
\subsection{Proof of Proposition~\ref{prop:sparse}}\label{pf:prop-sparse}
Let $x,x'\in\mathcal{A}$ and let $\mathcal{S}=supp(x)\cup supp(x')$. By assumption, we have $|supp(x)|\leq s$ and $|supp(x')|\leq s$. By union bound, we have $|\mathcal{S}|\leq 2s$. It follows that
\begin{align*}
\left|\langle Qx,Qx'\rangle-\langle x,x'\rangle\right| &= \left|\langle Q_\mathcal{S}x_\mathcal{S},Q_\mathcal{S}x'_\mathcal{S}\rangle-\langle x_\mathcal{S},x'_\mathcal{S}\rangle\right|\\
&= \left|\langle Q_\mathcal{S}^tQ_\mathcal{S}x_\mathcal{S},x'_\mathcal{S}\rangle-\langle x_\mathcal{S},x'_\mathcal{S}\rangle\right|\\
&= \left|\langle Q_\mathcal{S}^tQ_\mathcal{S}x_\mathcal{S}-x_\mathcal{S},x'_\mathcal{S}\rangle\right|\\
&= \left|\langle (Q_\mathcal{S}^tQ_\mathcal{S}-I_{2s})x_\mathcal{S},x'_\mathcal{S}\rangle\right|\\
& \leq \|(Q_\mathcal{S}^tQ_\mathcal{S}-I_{2s})x_\mathcal{S}\|\|x'_\mathcal{S}\|\\
& \leq \|Q_\mathcal{S}^tQ_\mathcal{S}-I_{2s}\|_{2\rightarrow 2}\|x_\mathcal{S}\|\|x'_\mathcal{S}\|\\
& \leq \delta_{2s}\|x_\mathcal{S}\|\|x'_\mathcal{S}\|\\
& = \delta_{2s}\|x\|\|x'\|\\
& \leq \delta_{2s}R^2.
\end{align*}
for any $x,x'\in\mathcal{A}$. We thus conclude that the matrix $Q$ is $\delta_{2s}R^2$-inner-product-preserving over $\mathcal{A}$.

\subsection{Proof of Lemma \ref{lm:support}}
{\em Proof:} To show that $\|x\|\leq R$ for any $(x,y)\in\mathcal{Z}$, suppose by contradiction that there exists $(x,y)\in\mathcal{Z}$ such that $\|x\|=\gamma > R$. By the continuity of the Euclidean norm $\|\cdot\|$, there exists a $\delta>0$ such that $\left|\|x'\|-\|x\|\right|<\gamma-R$ for all $x'\in \mathbb{R}^n$ such that $|x'-x|<\delta$. Let $\mathcal{A}:=\{x'\in \mathbb{R}^n: \; |x'-x|<\delta\}$. Note that $\mathcal{A}$ is open and $x\in \mathcal{A}$. By the definition of $\mathcal{Z}$, we have $\mu(\mathcal{A}\times\{y\})>0$. Further note that we have 
$$\|x'\|= \|x\|-(\|x\|-\|x'\|)=\gamma-(\|x\|-\|x'\|) > \gamma-(\gamma-R)=R$$ 
for any $x'\in\mathcal{A}$. It follows that
$$\mathbb{P}_{(\mathsf{x},\mathsf{y})\sim \mu}\left[\|\mathsf{x}\|> R\right] \geq \mu(\mathcal{A}\times\{y\})>0,$$
contradicting the assumption that $\mathbb{P}_{(\mathsf{x},\mathsf{y})\sim \mu}\left[\|\mathsf{x}\|\leq R\right]=1$. Therefore, we must have $\|x\|\leq R$ for any $(x,y)\in\mathcal{Z}$.

The fact that $\|x\|_0\leq s$ and $y\left(\langle w_0,x\rangle+b_0\right)\geq 1$ for any $(x,y)\in\mathcal{Z}$ can be proved similarly. The details are omitted from the paper. \hfill $\blacksquare$

\subsection{Proof of Lemma \ref{lm:strict-separation}}

{\em Proof:} For any $x\in co\left(\mathcal{X}^+\right)$, by Carath\'{e}odory's theorem  we can write $x=\sum_{i=1}^{n+1}\alpha_ix_i$ for some $x_i \in \mathcal{X}^+$, $i\in[n+1]$, and $\alpha_i \geq 0$, $i\in[n+1]$, such that $\sum_{i=1}^{n+1}\alpha_i=1$. By Lemma~1, we have
$$\langle w_0,x_i\rangle+b_0\geq 1$$
for any $i\in[n+1]$. It follows that
\begin{align*}
\langle w_0,x\rangle+b_0=\left\langle w_0,\sum_{i=1}^{n+1}\alpha_ix_i\right\rangle+b_0=\sum_{i=1}^{n+1}\alpha_i\left(\langle w_0,x\rangle+b_0\right)\geq \sum_{i=1}^{n+1}\alpha_i=1.
\end{align*}
Similarly, it can shown that for any $x\in co\left(\mathcal{X}^-\right)$, we have $\langle w_0,x\rangle+b_0\leq -1$. It follows that for any $x'\in\mathcal{X}^+$ and $x''\in\mathcal{X}^-$, we have
\begin{align*}
\langle w_0,x'-x''\rangle=\left(\langle w_0,x'\rangle+b_0\right)-\left(\langle w_0,x''\rangle+b_0\right) \geq 2
\end{align*}
and hence
\begin{align*}
\|x'-x''\| \geq \frac{\left|\langle w_0,x'-x''\rangle\right|}{\|w_0\|}\geq \frac{2}{\|w_0\|}.
\end{align*}
Taking infimum on both sides over $x'\in\mathcal{X}^+$ and $x''\in\mathcal{X}^-$ proves that $\delta \geq\frac{2}{\|w_0\|}>0$. \hfill $\blacksquare$

\subsection{Proof of Lemma \ref{lm:achieve-min-sep}}
{\em Proof:} By the definition of infimum, for each $n\in\mathbb{N}$ there exist $x'_n\in co\left(\mathcal{X}^+\right)$ and $x''_n\in co\left(\mathcal{X}^-\right)$ such that
$$\delta \leq \|x'_n-x''_n\|\leq \delta+\frac{1}{n}.$$
Since $co\left(\mathcal{X}^+\right)$ is compact, there exists a subsequence $(x'_{n_k}:k\in\mathbb{N})$ in $co\left(\mathcal{X}^+\right)$ and an element $\bar{x}'\in cl\left(co\left(\mathcal{X}^+\right)\right)$ such that
$$\lim_{k\in \infty}\|x'_{n_k}-\bar{x}'\|=0.$$
Similarly, there exists a subsequence $(x''_{n_k}:k\in\mathbb{N})$ in $co\left(\mathcal{X}^-\right)$ and an element $\bar{x}''\in cl\left(co\left(\mathcal{X}^-\right)\right)$ such that
$$\lim_{k\in \infty}\|x''_{n_k}-\bar{x}''\|=0.$$

Note that for any $k\in\mathbb{N}$, we have
\begin{align*}
\|\bar{x}'-\bar{x}''\| & = \|(x'_{n_k}-x''_{n_k})-(x'_{n_k}-x''_{n_k}+\bar{x}''-\bar{x}')\|\\
& \geq \|x'_{n_k}-x''_{n_k}\|-\|x'_{n_k}-x''_{n_k}+\bar{x}''-\bar{x}'\|\\
& =  \|x'_{n_k}-x''_{n_k}\|-\|(x'_{n_k}-\bar{x}')+(\bar{x}''-x''_{n_k})\|\\
& \geq \delta-\left(\|x'_{n_k}-\bar{x}'\|+\|\bar{x}''-x''_{n_k}\|\right).
\end{align*}
Let $k \rightarrow \infty$ and we have $\|\bar{x}'-\bar{x}''\| \geq \delta$. 

On the other hand, for any $k\in\mathbb{N}$ we have
\begin{align*}
\|\bar{x}'-\bar{x}''\| & = \|(x'_{n_k}-x''_{n_k})+(x''_{n_k}-x'_{n_k}+\bar{x}'-\bar{x}'')\|\\
& \leq \|x'_{n_k}-x''_{n_k}\|+\|x''_{n_k}-x'_{n_k}+\bar{x}'-\bar{x}''\|\\
& =  \|x'_{n_k}-x''_{n_k}\|+\|(x''_{n_k}-\bar{x}'')+(\bar{x}'-x'_{n_k})\|\\
& \leq \left(\delta+\frac{1}{n_k}\right)+\left(\|x''_{n_k}-\bar{x}''\|+\|\bar{x}'-x'_{n_k}\|\right).
\end{align*}
Let $k \rightarrow \infty$ and we have $\|\bar{x}'-\bar{x}''\| \leq \delta$. Combining the facts that $\|\bar{x}'-\bar{x}''\| \geq \delta$ and $\|\bar{x}'-\bar{x}''\| \leq \delta$ completes the proof that $\|\bar{x}'-\bar{x}''\| = \delta$. \hfill $\blacksquare$

\end{document}